\documentclass[journal]{IEEEtran}
\IEEEoverridecommandlockouts
\usepackage{cite}
\usepackage{amsmath,amssymb,amsfonts}
\usepackage{algorithmic}
\usepackage{graphicx}
\usepackage{textcomp}
\usepackage{multirow}
\usepackage{xcolor}
\usepackage[flushleft]{threeparttable}
\usepackage[a4paper, total={184mm,239mm}]{geometry}
\usepackage[normalem]{ulem}

\def\BibTeX{{\rm B\kern-.05em{\sc i\kern-.025em b}\kern-.08em
		T\kern-.1667em\lower.7ex\hbox{E}\kern-.125emX}}

\begin{document}

	\title{AHAR: Adaptive CNN for Energy-efficient Human Activity Recognition in Low-power Edge Devices\\
	}
	
	\author{Nafiul~Rashid,~\IEEEmembership{Student Member,~IEEE,}
		Berken Utku Demirel,
		and~Mohammad~Abdullah~Al~Faruque,~\IEEEmembership{Senior~Member,~IEEE}
		\thanks{ All the authors are with the Department of Electrical Engineering
			and Computer Science, University of California, Irvine, CA
			92697, USA, e-mail: (nafiulr@uci.edu)}
		\thanks{ Copyright (c) 2022 IEEE. Personal use of this material is permitted. However, permission to use this material for any other purposes must be obtained from the IEEE by sending a request to pubs-permissions@ieee.org.}
	}
	
	\markboth{IEEE Internet of Things Journal,~Vol.~XX, No.~XX, XXXX~2022}%
	{Shell \MakeLowercase{\textit{et al.}}: Bare Demo of IEEEtran.cls for IEEE Journals}
	
	\maketitle
	
	\begin{abstract}
		Human Activity Recognition (HAR) is one of the key applications of health monitoring that requires continuous use of wearable devices to track daily activities. This paper proposes an Adaptive CNN for energy-efficient HAR (AHAR) suitable for low-power edge devices. Unlike traditional adaptive (early-exit) architecture that makes the early-exit decision based on classification confidence, AHAR proposes a novel adaptive architecture that uses an output block predictor to select a portion of the baseline architecture to use during the inference phase. Experimental results show that traditional adaptive architectures suffer from performance loss whereas our adaptive architecture provides similar or better performance as the baseline one while being energy-efficient. We validate our methodology in classifying locomotion activities from two datasets- Opportunity and w-HAR. Compared to the fog/cloud computing approaches for the Opportunity dataset, our baseline and adaptive architecture shows a comparable weighted F1 score of 91.79\%, and 91.57\%, respectively. For the w-HAR dataset, our baseline and adaptive architecture outperforms the state-of-the-art works with a weighted F1 score of 97.55\%, and 97.64\%, respectively. Evaluation on real hardware shows that our baseline architecture is significantly energy-efficient (422.38x less) and memory-efficient (14.29x less) compared to the works on the Opportunity dataset. For the w-HAR dataset, our baseline architecture requires 2.04x less energy and 2.18x less memory compared to the state-of-the-art work. Moreover, experimental results show that our adaptive architecture is 12.32\% (Opportunity) and 11.14\% (w-HAR) energy-efficient than our baseline while providing similar (Opportunity) or better (w-HAR) performance with no significant memory overhead.
	\end{abstract}

	
	\section{Introduction}
	
\begin{figure}[t]
		\centering
		\includegraphics[trim={0cm .5cm 0cm 0cm},width=\linewidth]{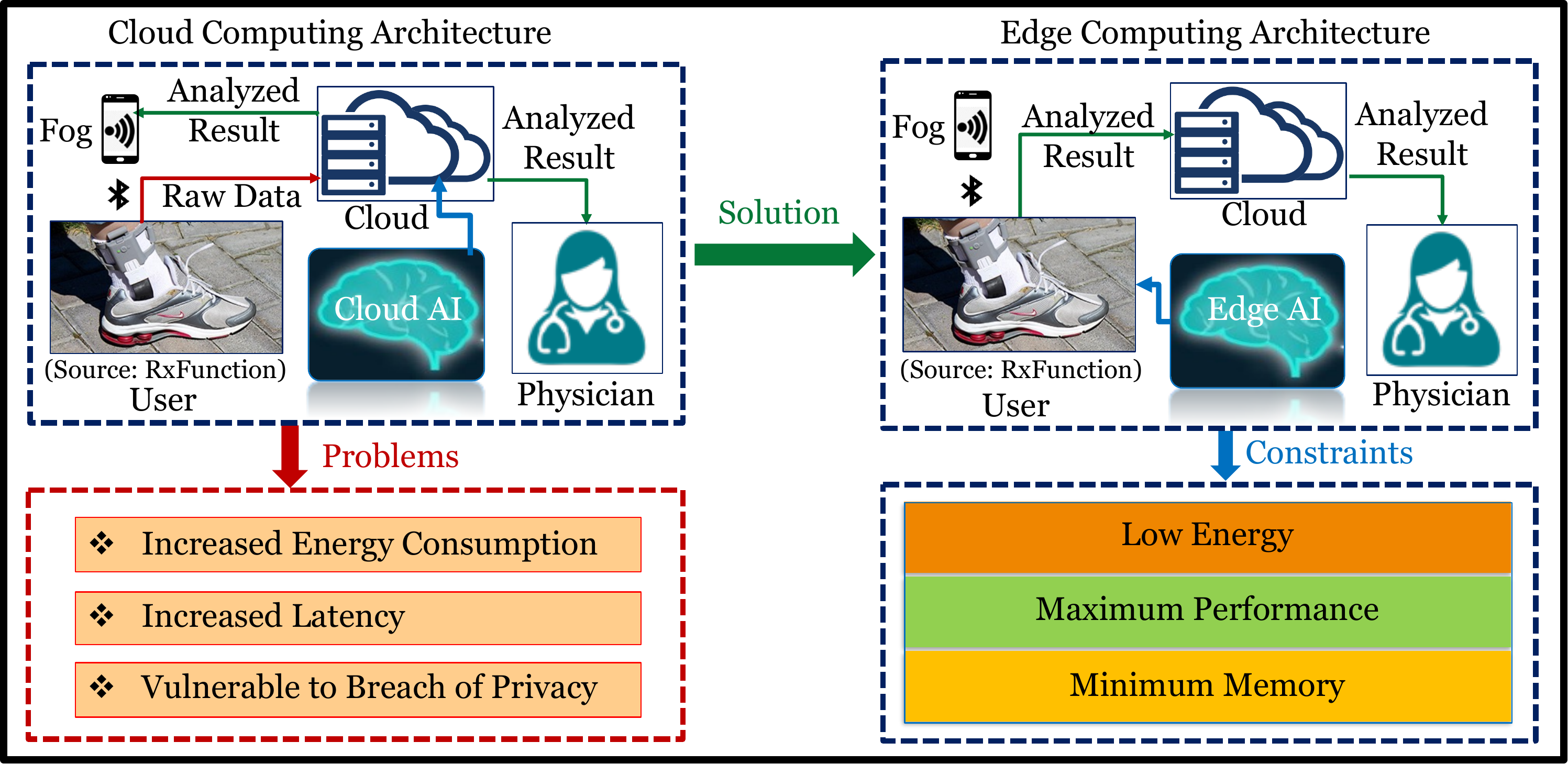}
		\caption{Shift from cloud computing to edge computing architecture}
		\label{Cloud2Edge}
	\end{figure}

	Human Activity Recognition (HAR) applications are useful tools for health monitoring, fitness tracking, and patient rehabilitation \cite{bort2014measuring, bourke2008threshold, parkinson_monitoring}. Since the HAR applications need continuous sensor data to infer user activity, advances in sensor technology \cite{wireless_manik} have enabled wide adoption of HAR applications in daily life. Smartphones have been significantly used for HAR in the past decade \cite{HHAR, smartphone_survey, lara2012mobile}. However, this kind of solution requires the user to continuously carry the phone which causes inconvenience. Moreover, the smartphone solutions consume higher energy in the range of watts \cite{smartphone_energy} which may hinder the primary use of the phones reducing the battery life. 
	Therefore, wearable devices have gained much popularity for HAR applications \cite{wearable_survey}. Moreover, the use of wearable devices enable remote monitoring of patients suffering from critical diseases like movement disorders in Parkinson's disease \cite{parkinson_monitoring}. However, most of the solutions \cite{khan2010triaxial, RF-4086Features, RF-SVM-KNN, Deep-CNN, Deep-CNN2, Deep-CNN3, DeepConvLSTM, Emilio-like-Survey, I-CNN} using wearable devices follow a fog/cloud computing approach as shown in Figure \ref{Cloud2Edge}. The collected data from wearable devices are sent over Bluetooth to a mobile phone (fog) \cite{HEAR} or remote server (cloud) where all the processing and classification takes place.  The daily use of these devices generates vast amounts of raw data, and sending them over Bluetooth entails higher energy consumption \cite{wearable_survey}. Additionally, it also introduces latency, which is unsuitable for real-time monitoring. Moreover, passing the raw data to a mobile phone makes the users' data vulnerable to privacy breaches. Many researchers \cite{Hierarchy_Henkel} followed a hierarchical approach where some simple activities are classified on the device where complex ones are transmitted over to the fog/cloud. Although this kind of solution saves computational energy to some extent, it still suffers from latency and privacy issues. Consequently, researchers shifted to an alternative architecture to overcome these limitations, which is called `edge computing' \cite{EDGE}, where all the processing is done on the device itself \cite{Umit_HAR, EMBC_2020}. Therefore, it reduces the energy consumption, latency, and vulnerability of privacy breaches. Figure \ref{Cloud2Edge} illustrates the shift from cloud to edge computing architecture.

	\begin{table*}[t]
		\centering
		\caption{Difference between Baseline and Adaptive Architecture for HAR Dataset}
		\label{layer_wise_table}
		\begin{tabular}{|c|c|c|c|c|c|c|c|}
			\hline
			\textbf{Architecture} & \textbf{Output} &  \textbf{Percentage of} & \textbf{Number of} & \textbf{Correct} & \textbf{Total FLOP} & \textbf{Total exec.} & \textbf{Total}\\
			\textbf{used} & \textbf{block used} &  \textbf{total segments} & \textbf{total segments} & \textbf{classification(\%)} & \textbf{count} & \textbf{time (ms)} & \textbf{energy ($\mu$J)}\\
			\hline\hline
			\textbf{Baseline} & \textbf{Second} & \textbf{100} & \textbf{4740} & \textbf{97.60} &\textbf{35,905,500} & \textbf{152,011.80} & \textbf{2,316,627.60}\\
			\hline\hline
			\multirow{3}{*}{\textbf{Adaptive}}  & First & 97.13 & 4604 & 95.06 & 26,606,516 & 121,361.44 & 1,849,564.92\\
			\cline{2-8}
			& Second & 2.87 & 136 & 2.87 & 1,030,200 & 4,361.52 & 66,468.64\\
			\cline{2-8}
			& \textbf{Overall} & \textbf{100} & \textbf{4740} & \textbf{97.93} & \textbf{27,636,716} & \textbf{125,722.96} & \textbf{1,916,033.56}\\
			\hline\hline
			\multicolumn{5}{|c|}{\textbf{Theoretical total saving due to adaptive architecture = Baseline - Adaptive}}  & \textbf{8,268,784} & \textbf{26,288.84} & \textbf{400,594.04}\\
			\hline
			\multicolumn{5}{|c|}{\textbf{Theoretical average saving per segment due to adaptive architecture}}  & \textbf{1,744.47} & \textbf{5.55} & \textbf{84.51}\\
			\hline
			
		\end{tabular}
	\end{table*}

	\begin{figure}[t]
	\centering
	\includegraphics[trim={15cm 21cm 11.5cm 3cm},clip, width=\linewidth]{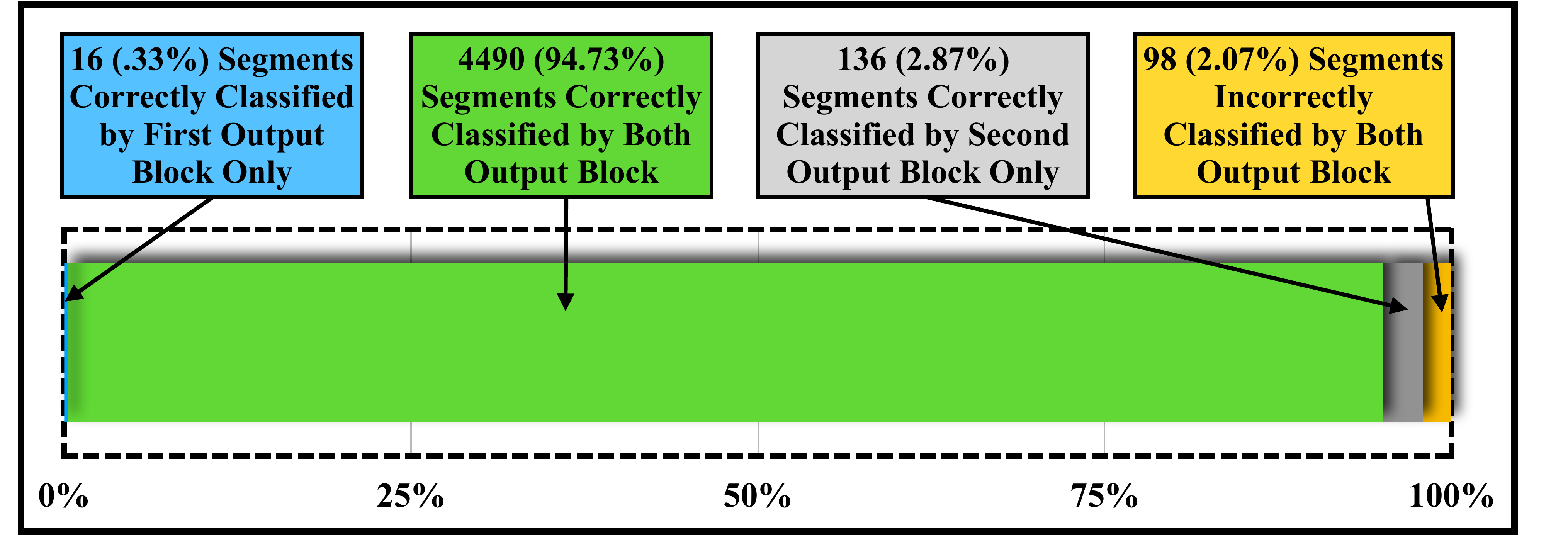}
	\caption{Blockwise multi-output CNN architecture performance breakdown}
	\label{fig:Venn_Motivation}
\end{figure}
	
	The small form factor of wearable devices imposes three constraints on the processing algorithms as shown in Figure \ref{Cloud2Edge}. The algorithms should consume low-energy, execute with minimum-memory, and provide maximum performance within the previous two constraints. State-of-the-art works on HAR are mostly intended for fog/cloud platform where they use complex machine learning \cite{khan2010triaxial, RF-4086Features, RF-SVM-KNN} and deep learning algorithms \cite{Deep-CNN, Deep-CNN2, Deep-CNN3, DeepConvLSTM, Emilio-like-Survey, I-CNN} to achieve high performance. They prioritize performance over the other two constraints, hence are not wearable device compatible. Machine learning algorithms perform classification based on the extracted features from the data which is often time and energy consuming, whereas, wearable device solutions should be fast and energy-efficient. Deep learning algorithms using Convolutional Neural Networks (CNN) \cite{CNN, rashid2021feature} have an advantage in this regard as they automatically extract features through convolution and do not require manual feature engineering or extraction. However, such deep networks require higher energy, memory, and execution time as they use a large number of layers. Therefore, for wearable device solutions CNN should be designed in such a way that satisfies the energy and memory constraints while maintaining reasonable performance. As CNN works in layers, it provides the flexibility to design a network by adding or removing layers as necessary in the training phase which is used to classify data during the inference phase. However, the full architecture from the training phase may not be needed at the inference phase as many of the data may be correctly classified using only the first few layers of the architecture. Therefore, if we use a portion of the network as needed, it will help to avoid redundant operations of the CNN architecture leading to energy efficiency while maintaining the performance. This technique is called adaptive (early-exit) or Conditional Deep Learning Network (CDLN) architecture and was adopted by many researchers \cite{panda2016conditional, scardapane2020should} for image classification or computer vision applications. The traditional adaptive architectures or CDLN makes the early-exit decision based on the classification confidence at each output (exit) layer. If the classification confidence of an output layer for a particular class exceeds a threshold they exit the network. However, such architectures may suffer from performance loss than the baseline architecture when the earlier layer misclassifies a segment with higher confidence which is demonstrated later in Table \ref{CDLN_performancetable}. Therefore, implementing adaptive architecture based on classification confidence does not ensure similar performance as the baseline. This motivates us to propose an adaptive architecture that uses an output block predictor to make the early-exit decision which will ensure similar or better performance as the baseline while providing energy efficiency. Sections \ref{Motivational_Example} and \ref{Observation} provide a motivational example along with the observation to support our proposed adaptive architecture.

	\begin{figure}[t]
		\centering
		\includegraphics[trim={11.7cm 14.4cm 7cm 7cm},clip, width=\linewidth]{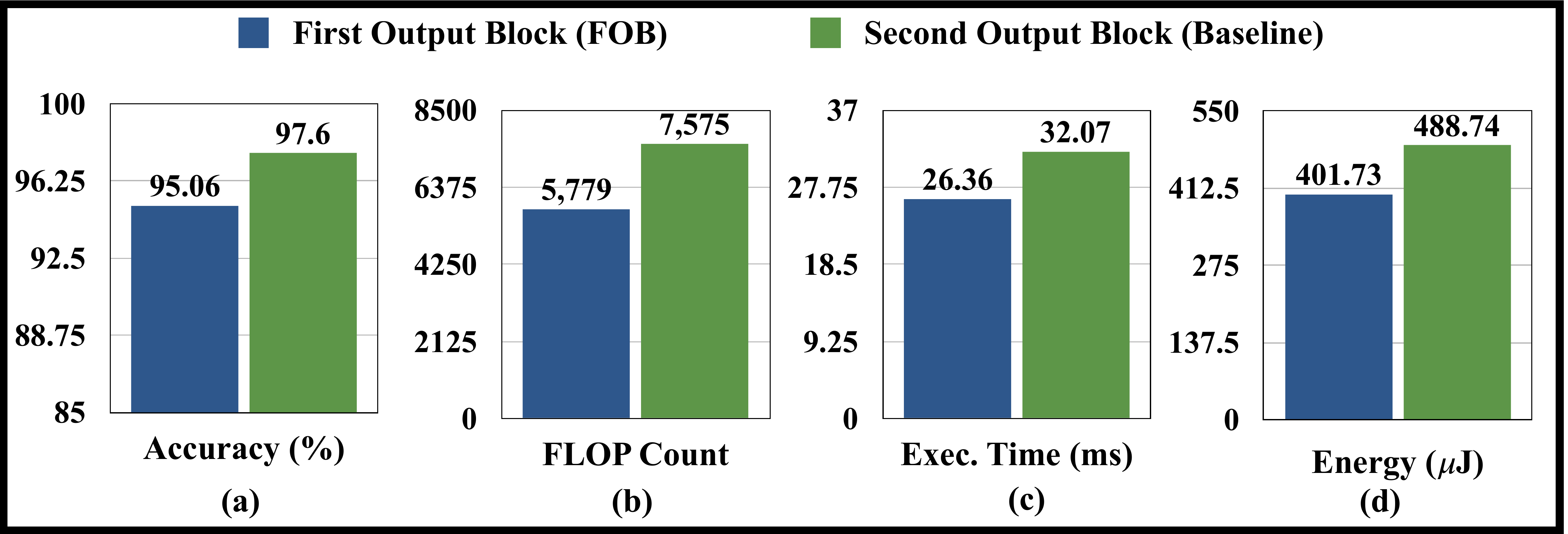}
		\caption{Blockwise statistics of multi-output CNN architecture}
		\label{fig:motivation_stats}
	\end{figure}
	
	\subsection{Motivational Example}
	\label{Motivational_Example}
	To demonstrate the advantage of an adaptive architecture we have conducted a small experiment. We have created a multi-output CNN architecture with 2 convolution blocks and 2 output blocks. One output block is used after each of the convolution blocks so that we can exit the architecture after any convolution block at the inference phase. The first convolution block consists of one convolution layer, one pooling layer, and one batch normalization layer. The second convolution block consists of one convolution and one batch normalization layer only. The output blocks contain either one or two dense layers which represents the output layer. The details of the multi-output CNN architecture is provided in Section \ref{sec:baseline_multioutput}. Throughout the rest of the paper, the first output block (FOB) is used as the portion of the CNN model that uses the first convolution block.  The second output block is referred to as the CNN architecture that uses two blocks of convolution which is the baseline architecture. We performed a 5-fold cross-validation of the multi-output CNN architecture with 4740 activity segments from the w-HAR dataset \cite{w_HAR}. Figure \ref{fig:Venn_Motivation} shows the Venn diagram for the multi-output CNN architecture performance where 94.73\% are correctly classified by both the FOB and baseline architecture. Only 0.33\% and 2.87\% of the segments are correctly classified by the FOB and baseline architecture respectively. Rest 2.07\% segments are incorrectly classified by both of them. Figure \ref{fig:motivation_stats}a shows the blockwise performance breakdown.  We find that the accuracy of the multi-output CNN architecture after the FOB, and baseline architecture are 95.06\%, and 97.60\%, respectively. Figure \ref{fig:motivation_stats}b shows the corresponding number of Floating Point Operations (FLOP) necessary to classify one activity segment after the FOB, and baseline architecture which are 5,799, and 7,575, respectively. Figure \ref{fig:motivation_stats}c shows the amount of execution time required to classify one activity segment on target wearable platform after FOB, and baseline architecture which are 26.36 $\mu$J, and 32.07 $\mu$J, respectively. Figure \ref{fig:motivation_stats}d demonstrates the amount of energy required to classify one activity segment on target wearable platform after FOB, and baseline architecture which are 401.73 $\mu$J, and 488.74 $\mu$J, respectively.
	
	\subsection{Observation and Problem Statement}
	\label{Observation}
	Figure \ref{fig:motivation_stats} demonstrates that the FLOP counts, execution time, and energy increases as performance increases from the first to second output block. To get a better performance, one would choose the second output block as the baseline architecture (as in our case) at the cost of increased energy. However, Figure \ref{fig:Venn_Motivation} shows that 94.73\% (4490) segments that are correctly classified by the baseline architecture are also correctly classified by the FOB. Therefore, using the baseline architecture for those segments would be redundant. If we can avoid these redundant operations, we can easily save some inference time and energy of the wearable devices. Therefore, instead of using a fixed baseline architecture, it would be energy-efficient if we could adaptively decide at the inference phase up to which output block we should use. As shown in Figure \ref{fig:Venn_Motivation}, if we could adaptively use the FOB to classify the 95.06\% (4490+16=4506) segments and use the baseline only for the 2.87\% (136) segments, overall accuracy (97.93\%) would be greater than that of the baseline architecture (97.60\%) at a much lower energy consumption. Table \ref{layer_wise_table} shows the theoretical breakdown of the performance, FLOP counts, execution time and energy of the adaptive architecture considering the FOB is also used for the 2.07\% (98) segments those are misclassified by both output block. Table \ref{layer_wise_table} demonstrates that using adaptive architecture, theoretically we can save a total of 8,268,784 FLOPs, 26,288.84 ms of execution time and 400,594.04 $\mu J$ of energy for 4740 segments. On average for each segment, we can save 1,744.47 FLOPs, 5.55 ms of execution time, and 84.51 $\mu J$ of energy using an adaptive architecture compared to the baseline architecture. In summary, an adaptive architecture would provide a much more energy-efficient solution than a baseline architecture while providing better or similar performance that is suitable for low-power wearable edge devices. On the other hand, traditional adaptive architectures or CDLN suffer from performance loss as the earlier layer misclassifies a segment with higher confidence and ends up exiting the network wrongly. As shown in Table \ref{CDLN_performancetable}, the performance of CDLN for various confidence thresholds at FOB. The maximum performance of CDLN is achieved for the confidence threshold of 0.9 which is still much less than our baseline architecture. Therefore, our adaptive architecture uses an output block predictor (instead of classification confidence) to make the early-exit decision.  Table \ref{CDLN_performancetable} shows that our adaptive architecture not only outperforms the traditional CDLN but also the baseline architecture for all performance metrics. It shows the efficacy of our adaptive architecture over traditional CDLN.
	
		\begin{table}[t]
		\centering
		\caption{Performance of CDLN for different FOB confidence threshold}
		\label{CDLN_performancetable}
		\begin{tabular}{|c|c|c|c|c|}
			\hline
			\textbf{Method} & \textbf{Weighted F1} & \textbf{Accuracy} & \textbf{Precision} & \textbf{Recall} \\
			\hline\hline
			\textbf{CDLN (th = 0.5)} & 94.49 & 95.06 & 94.87  &  95.05 \\
			\hline
			\textbf{CDLN (th = 0.6)} & 94.71  & 95.25 & 95.09 &  95.24 \\
			\hline
			\textbf{CDLN (th = 0.7)} & 94.93 & 95.42 & 95.31 & 95.41  \\
			\hline
			\textbf{CDLN (th = 0.8)} & 95.01 & 95.48 &  95.37 &  95.47 \\
			\hline
			\textbf{CDLN (th = 0.9)} & 95.23 & 95.68 &  95.59 & 95.67\\
			\hline
			\textbf{Baseline [Ours]} & \textbf{97.55} & \textbf{97.60}&  \textbf{97.57} & \textbf{97.60}\\
			\hline
			\textbf{Adaptive [Ours]} & \textbf{97.64} & \textbf{97.70} &  \textbf{97.69} & \textbf{97.70}\\
			\hline
		\end{tabular}
	\end{table}

	\subsection{Novel Contributions}
	The novel contributions of this paper are as follows:
	\begin{itemize}

		\item A novel Adaptive CNN architecture for HAR (AHAR) that uses an output block predictor to select a portion of the baseline architecture as needed during the inference phase. To the best of our knowledge, we are the first to investigate such an adaptive CNN architecture for HAR application.
		\item Evaluation of our methodology in classifying locomotion activities from Opportunity \cite{OPPORTUNITY} and w-HAR \cite{w_HAR} dataset. In comparison to the fog/cloud computing approaches on the Opportunity dataset, both our baseline and adaptive architecture shows a comparable weighted F1 score of 91.79\%, 91.57\% respectively. For the w-HAR dataset, both our baseline and adaptive architecture outperforms the state-of-art-work with a weighted F1 score of 97.55\% and 97.64\%, respectively.
		\item Evaluation on real hardware shows that our baseline architecture is significantly energy-efficient (422.38x less) and memory-efficient (14.29x less) compared to the works on Opportunity dataset. For w-HAR dataset, our baseline architecture requires 2.04x less energy and 2.18x less memory compared to the state-of-the-art work on wearable devices.
		\item Experimental validation show that our adaptive architecture is 12.32\% (Opportunity) and 11.14\% (w-HAR) energy-efficient than our baseline while providing similar (Opportunity) or better (w-HAR) performance with no significant memory overhead.
	\end{itemize}

	\section{Related Works}	
	\subsection{Works on Human Activity Recognition}	
	The main goal of our paper is to propose a wearable device solution for classifying locomotion activities. Therefore, to validate our proposed methodology, we have considered the Opportunity \cite{OPPORTUNITY} and w-HAR \cite{w_HAR} datasets that has labeled locomotion data from wearable devices. Accordingly, we will discuss and compare against the works mentioned in Table \ref{Methodology_wise_table} that have used either of these two datasets for classifying the locomotion activities.
	
	\begin{table}[t]
		\centering
		\begin{threeparttable}
		\caption{Summary of Related Works}
		\label{Methodology_wise_table}
		\begin{tabular}{|c|c|c|c|c|c|}
			\hline
			\multirow{2}{*}{\textbf{Work} } & \textbf{Data} &{\textbf{\# of}} & \textbf{Classifier} & \multirow{2}{*}{\textbf{Adaptive}} &\textbf{Computing}\\
			& \textbf{used} &\textbf{chan.}  & \textbf{used} & &\textbf{platform}\\
			\hline\hline
			\cite{RF-4086Features} & Opp. & 117 & RF (n=[40,95]) & No &Fog/Cloud\\
			\hline
			\cite{DeepConvLSTM} &  Opp. & 113 & CNN, LSTM & No & Fog/Cloud\\
			\hline
			\cite{Emilio-like-Survey} & Opp.  & 6 & 2-D CNN & No & Fog/Cloud\\
			\hline
			\cite{I-CNN} & Opp.  &  113 & Deep CNN & No & Fog/Cloud\\
			\hline
			\cite{Umit_HAR}  & w-HAR & 4 & SVM, DT, NN & No & Edge\\
			\hline
			\textbf{Ours} & \textbf{Both} & \textbf{7} & \textbf{DT, 1-D CNN} & \textbf{Yes} &\textbf{Edge}\\
			\hline
		\end{tabular}
		\begin{tablenotes}
				\item \textit{Opportunity (Opp.)}
			\end{tablenotes}
		\end{threeparttable}
	\end{table}
	
	As shown in Table \ref{Methodology_wise_table}, works \cite{RF-4086Features, DeepConvLSTM, Emilio-like-Survey, I-CNN} have used Opportunity dataset for classifying 4 locomotion activities - \textit{Stand, Walk, Lie, Sit}. In \cite{RF-4086Features} the authors proposed an activity-recognition algorithm based on the random forest classifier by extracting 4086 features which are from both time and frequency domain. They achieve a weighted F1 score of 90\%. Authors in \cite{DeepConvLSTM} use deep CNN architecture composed of 4 convolutional and 2 LSTM recurrent layers and achieves a weighted F1 score 93\%. The work in \cite{Emilio-like-Survey} achieves an weighted F1 score of 92.57\% using a two dimensional CNN architecture. Finally, the work in \cite{I-CNN} used a CNN architecture that combines temporal and spatial convolutions to extract appropriate features to make it suitable for mobile devices. Their solution achieves a weighted F1 score of 92.5\%.
	
	\begin{figure*}[t]
		\centering
		\includegraphics[width=\linewidth]{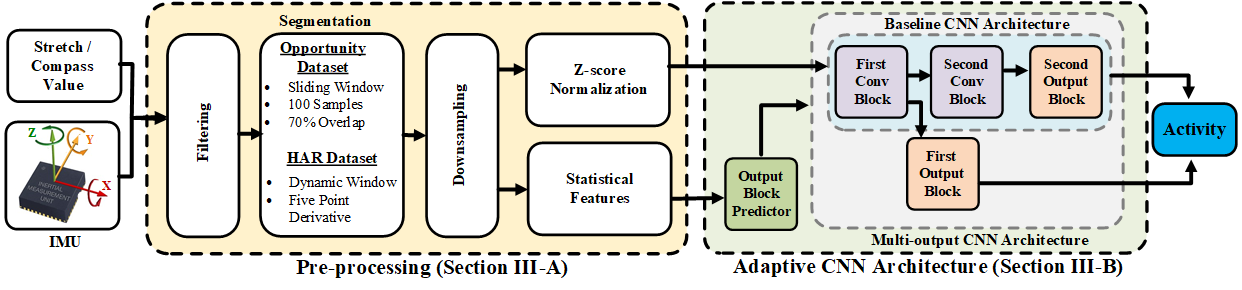}
		\caption{Overview of our proposed  AHAR methodology}
		\label{Methodology}
	\end{figure*}
	
	On the other hand, the work in \cite{Umit_HAR} used w-HAR dataset to propose a baseline and an activity-aware classifier for classifying 8 locomotion activities in wearable devices. The baseline and activity-aware classifier achieves an weighted F1 score of 94.96\% and 97.37\% respectively. The baseline architecture uses 120 statistical and frequency domain features whereas the activity-aware classifier works in hierarchical order. First, it classifies the activities as static (\textit{Lie down, Sit, Stand}) or dynamic (\textit{Jump, Walk, Stairs down, Stairs up, Transition}) by feeding 8 statistical features (mean, variance, minimum, maximum) to a support vector machine (SVM) classifier. Then, if the activity is classified as static, a decision tree is used to classify it further with the same statistical features. Otherwise, the other 112 frequency domain features (FFT) are extracted and together 120 features are fed to a neural network (NN) to classify dynamic activities. Table \ref{Methodology_wise_table} shows a summary of the related works.
	
	\subsection{Energy-efficient CNN Design Approaches}
	
	The deep architecture of CNN with hundreds of layers are very computationally expensive and not suitable for energy and memory constraint wearable devices. Therefore, different approaches have been introduced in the literature to make it energy and memory-efficient while maintaining similar or competitive performance. Such approaches can be broadly classified into two categories - 1) Software-based approach, 2) Hardware-based approach.
	
	The software-based approaches can be further classified into 2 phases - 1) Offline or training phase, 2) Online or inference phase.
	The software-based approaches in the training phase can be broadly divided into 3 types - a) Neural Architecture Search (NAS), b) Network Pruning, c) Model Compression. NAS looks for optimum network parameters from a search space using reinforcement learning \cite{lu2019neural} or gradient-based methods \cite{liu2018darts} or multi-objective bayesian optimization \cite{odema2021energy, EExNAS, odema2021lens}. Network pruning performs random pruning of a portion of the big network, retraining it, and repeating the process until it achieves the desired performance \cite{cai2020once}. Finally, model compression involves binarization \cite{courbariaux2016binarized} or quantization \cite{wang2019haq} of network weights to reduce the model size to make it memory-efficient. Another model compression technique is knowledge distillation where a smaller network (student model) is taught, step by step, exactly what to do using a bigger already trained network (teacher model) \cite{hinton2015distilling}. Regardless of the methods used, the final model from the training phase is considered as the baseline classifier to be used at the inference phase.
	
	Software-based approach designed for the inference phase is called adaptive (early-exit) or Conditional Deep Learning Network (CDLN) architecture \cite{panda2016conditional, scardapane2020should}. If the input data is classified with enough confidence after a convolutional layer then it considers that as the final class without further proceeding to the next layers of convolution. However, they may suffer from performance loss if the earlier layer misclassifies a segment with higher confidence and exits the network wrongly.

	It is to note that, the software-based approaches from the training and inference phase are independent of each other and they can be applied together as well. For example, during the inference phase, one can apply the early-exit mechanism to a baseline architecture that has been finalized at the training phase by using any of the NAS, network pruning, or model compression techniques.
	
	Hardware-based approaches usually focus on the design of custom hardware such as accelerators which are specifically designed for CNN \cite{chen2016eyeriss, chen2019eyeriss}. The main goal is to make the inference phase faster thereby making it more energy-efficient. 
	
	In this paper, we mainly focus on the inference phase of the software-based approach which allows early-exit. However, unlike the related works \cite{panda2016conditional, scardapane2020should}, we propose a novel adaptive CNN architecture that uses an output block predictor (instead of classification confidence) to make the early-exit decision without any performance loss. To the best of our knowledge, we are the first to investigate such an adaptive CNN architecture for HAR application.

	\section {Proposed Methodology}
	\subsection{Pre-processing Steps}
	
	\subsubsection{Filtering} As shown in Figure \ref{Methodology}, the pre-processing starts with the denoising and smoothing. Raw data is filtered using a moving average filter with a window of 8 samples to smoothen it. Then the filtered data is segmented. 
	
	\subsubsection{Segmentation} As the data from different datasets varies, we apply different segmentation technique for two datasets used in our paper. For the Opportunity \cite{OPPORTUNITY} dataset, the segmentation of filtered data is done using a sliding window of 100 samples with 70\% overlap. As the data is collected at a sampling rate of 30 Hz, each segment of data captures 3.33 seconds of data. For the w-HAR \cite{w_HAR} dataset, we follow the dynamic segmentation technique based on five-point derivative on the stretch sensor data as mentioned in \cite{Umit_HAR}. The details of the datasets are given in Section \ref{Datasets}. 
	\subsubsection{Downsampling}
	Once segmented, we downsample each segment to 32 samples. Downsampling helps in two ways - 1) Lower number of samples in a segment requires less computation for the CNN architecture which makes the solution energy-efficient. 2) Downsampling to a fixed number of samples also helps when we perform dynamic segmentation as CNN requires a fixed size for the input segments.

	\subsubsection{Calculating Statistical Features}
	\label{statistical_features}
	Next, we extract simple statistical features for each segment to be used by our output block predictor to implement our adaptive CNN architecture. The details of the output block predictor is given in Section \ref{Output_Block_Predictor}. We have used a minimum number of features to ensure minimal overhead for our adaptive architecture. For the segments in the Opportunity dataset, we extract 4 features (mean acceleration along X and Z axis, minimum and maximum value of angular velocity along Z axis). For the segments in w-HAR dataset, we extract 6 features (mean acceleration along X and Z axis, minimum and maximum of gyroscope value along Z axis, minimum and maximum of Stretch sensor value). These extracted features will be used by our output block predictor to decide which output block to be used at the inference phase to classify a particular segment.
	
	\subsubsection{Z-score Normalization}
	Before passing the downsampled segments to our multi-output CNN architecture, we normalize each segment using Z-score normalization (Eq. \ref{z_score}) to reduce the effect of any outlier samples in the corresponding segments.
	\begin{equation}
	\label{z_score}
	Z_{i} = \frac {X_{i}- \bar{X}}{S}
	\end{equation}
	For a particular segment, $Z_{i}$ is Z-score value of the $i^{th}$ sample $X_{i}$ whereas, $\bar{X}$ and  $S$ are the mean and standard deviation of the samples in that segment.
	
	\subsection{Adaptive CNN Architecture}
	Our designed adaptive CNN architecture consists of two parts - 1) Multi-output CNN architecture that classifies the segments of activity, 2) Output block predictor that decides which output block of the multi-output CNN architecture is to be used at inference phase based on some statistical features of each segment.
	\subsubsection{Multi-output CNN Architecture}
	\label{sec:baseline_multioutput}
\begin{figure}[t]
		\centering
		\includegraphics[width=\linewidth]{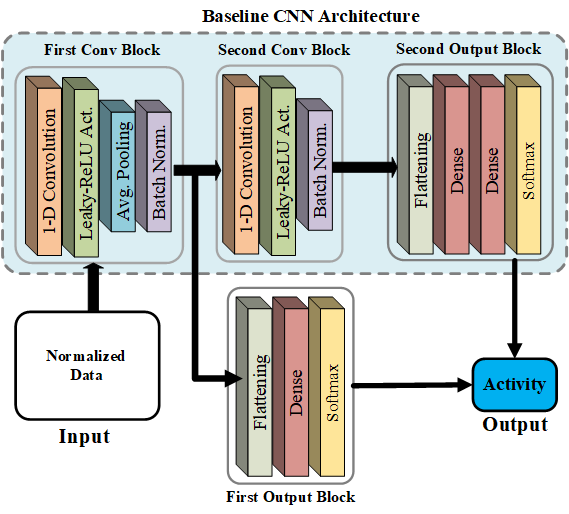}
		\caption{Multi-output CNN architecture layout}
		\label{Multi-output_CNN_Architecture_Layout}
	\end{figure}

	As our target platform is the low-power edge devices, we design the multi-output CNN architecture considering the resource constraints of the wearable devices. Our multi-output CNN architecture consists of 2 convolution blocks and 2 output blocks. Each convolution block is followed by one output block. Figure \ref{Multi-output_CNN_Architecture_Layout} shows the architecture layout of our multi-output CNN architecture. The first convolution block consists of one convolution layer which is passed through \textit{Leaky-ReLU} activation, one average-pooling layer, and one batch normalization layer whereas the second convolution block has one convolution layer which is passed through \textit{Leaky-ReLU} activation, and one batch normalization layer. The first output block consists of one flattening layer, and one dense layer which is passed through \textit{Softmax} activation. The second output block consists of one flattening layer, and 2 dense layers which are followed by the \textit{Softmax} activation as well. The details of the architecture parameters for each of the layers are given in Table \ref{Multi-output_CNN_Architecture_Details}. As shown in Table \ref{Multi-output_CNN_Architecture_Details}, the total number of parameters required to classify a segment after first output block (FOB) and second output block (baseline architecture) is 240+(31*$n_c$), and 744+(17*$n_c$), respectively where $n_c$ is the number of output classes. For the Opportunity dataset, we have 4 output classes and for the w-HAR dataset, we have 8 output classes.
	
	\begin{table}[t]
		\centering
		\begin{threeparttable}
			\caption{Multi-output CNN Architecture Details}
			\label{Multi-output_CNN_Architecture_Details}
			\begin{tabular}{|c|c|c|c|c|c|}
				\hline
				\textbf{Layer} & \textbf{Kernel} & \textbf{Stride} &  \textbf{Act.} & \textbf{Output} &  \textbf{\# of} \\
				\textbf{name} & \textbf{size} & \textbf{size} &  \textbf{func.}& \textbf{shape} &  \textbf{param.} \\
				\hline\hline
				Input & - & - & - & 32x7 & 0\\
				\hline
				Conv 1 & 5 & 3 & LR & 10x6 & 216\\
				\hline
				Pool 1 & 2 & 2 & - & 5x6 & 0\\
				\hline
				BN 1  & - & - & - & 5x6 & 24\\
				\hline
				Flat 1 & - & - & - & 30 & 0\\
				\hline
				Dense 1 & - & - & SM & $n_c$ & 31*$n_c$\\
				\hline
				Conv 2 & 4 & 1 & LR & 2x8 & 200\\
				\hline
				BN 2 & - & - & - & 2x8 & 32\\
				\hline
				Flat 2 & - & - & - & 16 & 0 \\
				\hline
				Dense 2 & - & - & LR & 16 & 272 \\
				\hline
				Dense 3 & - & - & SM & $n_c$ & 17*$n_c$ \\
				\hline
				\hline
				\multicolumn{5}{|c|}{\textbf{Number of parameters after FOB}} & \textbf{240+(31*\textbf{$n_c$})}\\ 
				\hline
				\multicolumn{5}{|c|}{\textbf{Number of parameters after baseline architecture}} & \textbf{744+(17*\textbf{$n_c$})}\\ 
				\hline
			\end{tabular}
			\begin{tablenotes}
				\item \textit{Batch Normalization (BN), Leaky-ReLU (LR), Softmax (SM)}
			\end{tablenotes}
		\end{threeparttable}
	\end{table}
	
	\subsubsection{Output Block Predictor (OBP)}
	\label{Output_Block_Predictor}
	The output block predictor (OBP) is very crucial for our adaptive CNN architecture as the performance of the adaptive architecture greatly depends on the OBP. The better the performance of the OBP is, the better the performance of our adaptive architecture will be. We use a decision tree as our OBP to decide which output block to be used at the inference phase based on the statistical features for each segment. Therefore, instead of using the baseline architecture (second output block) to classify all the segments, we will adaptively use FOB or baseline based on the decision of output block predictor. This will help to avoid unnecessary computation up to the second output block of baseline architecture as some of the segments might be correctly classified just after FOB. As the goal of our adaptive architecture is to ensure energy efficiency compared to the baseline architecture, the OBP should be designed in such a way that satisfies the following constraint:
	
	\begin{equation}
	\label{constraint_1}
	[N \times E_{pred} + N_1\times E_1+  N_2\times E_2 ] < [N \times E_2]
	\end{equation}
	
	Where $E_{pred}$, $E_1$, $E_2$ is the amount of energy - for the OBP and the FOB, and baseline architecture respectively. $N_1$, and  $N_2$ are the number of segments that are classified by the FOB, and baseline architecture, respectively where, $N_1+N_2=N$. Equations \ref{constraint_1}  ensures that the total amount of energy needed to classify N segments using adaptive architecture should be less than that of the baseline one.

	\section{Experimental Setup}
	\subsection{Datasets}
	\label{Datasets}
	\subsubsection{Opportunity Dataset \cite{OPPORTUNITY}}
	Opportunity dataset contains multimodal data from different wearable, object, and ambient sensors to benchmark the works on human activity recognition. The dataset contains a total of 6 hours of recording from 4 subjects. Each subject performs five sessions of Activities of Daily Living (ADL) and a drill session. The dataset is labeled for different gesture and locomotion activities. In our work, we use the locomotion activities (\textit{Stand, Walk, Sit, Lie down}) as our goal is to propose a wearable device solution that can classify the locomotion activities on the device itself. Therefore, we use only 7 channels of data in total where 3 channels (\textit{accX, accY, accZ}) are from accelerometer on the upper right knee and the other 4 channels (\textit{AngVelBodyFrameX, AngVelBodyFrameY, AngVelBodyFrameZ, Compass}) are from the Inertial Measurement Unit (IMU) on the right shoe. The channels are selected as they are suitable for designing a wearable device where the sensors are in close proximity while collecting maximum information with minimum channels. All the data are collected at 30 Hz from all the sensors.
	
	\subsubsection{w-HAR Dataset \cite{w_HAR}}
	w-HAR dataset contains wearable sensor data using IMU and stretch sensors from 22 subjects while performing 7 different locomotion activities (\textit{Jump, Lie down, Sit, Stairs down, Stairs up, Stand, Walk}). Additionally, they also labeled the \textit{Transition} between the activities. The dataset has 7 channels of data where 6 channels (\textit{Ax, Ay, Az, Gx, Gy, Gz}) are from the IMU on the right ankle and 1 channel (\textit{Stretch value}) is from the stretch sensor on the right knee. We use all 7 channels from this dataset as it is targeted towards wearable device design for locomotion activities. The IMU data is collected at 250 Hz and the stretch sensor data is collected at 25 Hz.
	
	\subsection{Training Multi-output CNN Classifier} 
	\label{Training_Multi-output_CNN_Classifier}
	As mentioned above, we train and test our multi-output CNN classifier on two different datasets. To ensure a fair comparison with the related works on locomotion activity recognition from the Opportunity dataset, we use similar distribution of training, testing and validation data as provided in the Opportunity challenge. Therefore, for training data we use - ADL1, ADL2, ADL3, ADL4, ADL5, DRILL data from subject 1; ADL1, ADL2, DRILL data from subject 2 and 3. The ADL3 data from subject 2 and 3 is used for validation. Finally, the classifier is tested on the ADL4 and ADL5 data from the subject 2 and 3. The classifier is trained for 100 epochs with \textit{Sparse Categorical Cross Entropy} as the loss function. \textit{Adam} optimizer is used to train the models with a learning rate of .007.
	
	For the w-HAR dataset, we perform a stratified 5-fold cross-validation as there is no specific distribution of train test data. Therefore, 80\% of the data is used for training, and the rest 20\% is used for testing. Moreover, 20\% of the training data is used for validation during training. For this dataset, the classifier is trained for 300 epochs with \textit{Sparse Categorical Cross Entropy} as the loss function. \textit{Adam} optimizer is used to train the models with a learning rate of 0.01.
	
	\begin{table}[t]		
		\centering
		\caption{Data Labeling Mechanism for Output Block Predictor}
		\label{OBP_Dataset_Labeling}
		\begin{tabular}{|c|c|c|c|}
			\hline
			\textbf{Cases} & \textbf{FOB} & \textbf{Baseline} & \textbf{Assigned label} \\
			\hline\hline
			\textbf{Both} & \checkmark & \checkmark & 1   \\
			\hline
			\textbf{FOB only} & \checkmark  & $\times$ & 1  \\
			\hline
			\textbf{Baseline only} & $\times$ & \checkmark &  2  \\
			\hline
			\textbf{None} & $\times$ & $\times$ &  1 \\
			\hline
		\end{tabular}
	\end{table} 
	
	\subsection{Training Output Block Predictor} 
	To train the output block predictor (OBP), we first generate a dataset based on the performance of the best multi-output CNN model for each of the Opportunity and w-HAR datasets. Then for each of the segments in the dataset, we determine which output block of the multi-output classifier can correctly classify them. For an activity segment, there are 4 different possible cases in our multi-output classifier as shown in Table \ref{OBP_Dataset_Labeling}.  If the segment is correctly classified by both output blocks we would want to use the FOB to save energy hence it is labeled as 1. If it is correctly classified by either FOB or baseline architecture only, it will be labeled as either 1 or 2 respectively. Finally, if it is misclassified by both FOB and baseline architecture that should also be labeled as 1 to avoid unnecessary computation by second output block to classify that segment.  Thus, the activity segments of each dataset are labeled which is used as the true label to train and test the OBP (decision tree).

	\begin{table*}[t]
		\centering
		\caption{Confusion Matrix of Different Output Blocks on Opportunity Dataset}
		\label{Opp_Confusion_Matrix}
		\begin{tabular}{|c||c|c|c|c||c|c|c|c||c|c|c|c|}
			\hline
			\multirow{2}{*}{\textbf{True label}} &
			\multicolumn{4}{c||}{\textbf{FOB}} &\multicolumn{4}{c||}{\textbf{Baseline}} & \multicolumn{4}{c|}{\textbf{Adaptive}} \\
			\cline{2-13}
			& \textbf{Stand} & \textbf{Walk} & \textbf{Lie} & \textbf{Sit} & \textbf{Stand} & \textbf{Walk} & \textbf{Lie} & \textbf{Sit}& \textbf{Stand} & \textbf{Walk} & \textbf{Lie} & \textbf{Sit}\\
			\hline\hline
			\textbf{Stand} & \textbf{1090} &  108 & 1 & 13 & \textbf{1103} & 98 &  1 & 10 &\textbf{1091} & 107 &   1  & 13\\
			\hline
			\textbf{Walk} & 103 & \textbf{813} & 0 & 7 & 107 &  \textbf{807} & 1 & 8 & 103 & \textbf{813} &   0  &  7\\
			\hline
			\textbf{Lie} & 0 & 117 & \textbf{63} & 5 & 0 & 0 & \textbf{184} & 1 & 0 & 1  & \textbf{181}  &  3\\
			\hline
			\textbf{Sit} & 18 & 6 &2 & \textbf{785} & 21 &  1 &  9 & \textbf{780} & 18 & 3  &  8 & \textbf{782}\\
			\hline
			
		\end{tabular}
	\end{table*}
	
	And the input to the OBP is the statistical features for each activity segment as calculated in Section \ref{statistical_features}. For the Opportunity dataset, we use 4 features whereas for the w-HAR dataset we use 6 features. To train and test the OBP for Opportunity dataset, we use the same training and testing segments as used in training and testing the multi-output CNN architecture as mentioned in Section \ref{Training_Multi-output_CNN_Classifier}. For OBP of the w-HAR dataset, we use stratified 5-fold cross-validation where 80\% data is used for training and 20\% data is used for testing.
	
	\begin{table}[t]		
		\centering
		\caption{Performance Comparison of Related Works on Opportunity Dataset for Locomotion (4 Activities)}
		\label{Opportunity_Performance_Table}
		\begin{tabular}{|c|c|c|c|c|}
			\hline
			\textbf{Works} & \textbf{Weighted F1} & \textbf{Accuracy} & \textbf{Precision} & \textbf{Recall} \\
			\hline\hline
			\textbf{RF\cite{RF-4086Features}} & 90.00 & -  & - &  - \\
			\hline
			\textbf{CNN,RNN\cite{DeepConvLSTM}} & 93.00 & - & - & - \\
			\hline
			\textbf{2-D CNN\cite{Emilio-like-Survey}} & 92.57 &  - & - & - \\
			\hline
			\textbf{1-D CNN\cite{I-CNN}} & 92.50 & - & - & - \\
			\hline
			\textbf{FOB [Ours]} & 87.24 & 87.86 & 88.54 & 87.86  \\ 
			\hline
			\textbf{Baseline [Ours]} &  91.79 & 91.79 & 91.80  & 91.79 \\
			\hline
			\textbf{Adaptive [Ours]} &  91.57 & 91.57 & 91.57  & 91.57 \\
			\hline
		\end{tabular}
	\end{table} 
	
	\begin{table}[t]		
		\centering
		\caption{Performance Comparison of Related Works on w-HAR Dataset for Locomotion (8 Activities)}
		\label{HAR_Performance_Table}
		\begin{tabular}{|c|c|c|c|c|}
			\hline
			\textbf{Works} & \textbf{Weighted F1} & \textbf{Accuracy} & \textbf{Precision} & \textbf{Recall} \\
			\hline\hline
			\textbf{Baseline\cite{Umit_HAR}} & 94.96 & 94.87 & 95.14  &  94.87 \\
			\hline
			\textbf{Activity-aware\cite{Umit_HAR}} & 97.37  & 97.34 & 97.45 &  97.34 \\
			\hline
			\textbf{FOB [Ours]} & 94.45 & 95.06 & 94.87 & 95.06  \\
			\hline
			\textbf{Baseline [Ours]} & 97.55 & 97.60 &  97.57 &  97.60 \\
			\hline
			\textbf{Adaptive [Ours]} & 97.64 & 97.70 &  97.69 & 97.70\\
			\hline
		\end{tabular}
	\end{table} 
	
	\subsection{Wearable Platform}
	\label{Wearable_Platform}
	Our proposed methodology is designed for low-power, low-memory wearable edge devices. Therefore, we evaluate our classifier on an ultra-low-power 32-bit microcontroller EFM32 Giant Gecko (EFM32GG-STK3700A) \cite{EFMGG} which has an ARM Cortex–M3 processor with a maximum clock rate of 48 MHz. It has 128 KB of RAM, 1 MB of Flash.

	\section{Experimental Results and Analysis}
	As the number of segments for different activities in both the datasets are highly imbalanced, only classification accuracy is not appropriate to measure performance. Therefore, to ensure proper performance evaluation, we use precision, recall, and weighted F1 score in addition to accuracy. The metrics used for evaluation are given below:
	\begin{equation}
	Accuracy = \frac {TP+TN}{TP+FP+TN+FN}
	\end{equation}
	
	\begin{equation}
	Precision = \frac {TP}{TP+FP}
	\end{equation}
	
	\begin{equation}
	Recall = \frac {TP}{TP+FN}
	\end{equation}
	
	\begin{equation}
	WF_1 =\sum_{i}^{n_c} 2* w_i \frac {Precision_i.Recall_i}{Precision_i+Recall_i}
	\end{equation}
	
	Where TP, TN, FP, FN represents True Positives, True Negatives, False Positives, and False Negatives respectively. The activity classes are indexed by \textit{i}, and $w_i$=$n_i/N$. $n_i$ is the number of activity segments in each class, and $N$ is the total number of activity segments.
	
	\begin{table}[t]
		\centering
		\caption{Confusion Matrix of First Output Block on w-HAR Dataset}
		\label{HAR_Confusion_Matrix_FOB}
		\begin{tabular}{|c||c|c|c|c|c|c|c|c|}
			\hline
			\multirow{2}{*}{\textbf{True label}} &
			\multicolumn{8}{c|}{\textbf{First Output Block (FOB)}} \\
			\cline{2-9}
			& \textbf{J} & \textbf{L} & \textbf{S} & \textbf{SD} & \textbf{SU} & \textbf{ST} & \textbf{W} & \textbf{T}\\
			\hline\hline
			\textbf{J} &\textbf{445}& 0 & 0 & 2 & 0 & 3 & 2 & 6 \\
			\hline
			\textbf{L} & 0  & \textbf{474}  &  0 &   0  &  0   & 0 &   0  &  0 \\
			\hline
			\textbf{S} &0  &  0  & \textbf{687}  &  0  &  0  &  9  &  0  &  0 \\
			\hline
			\textbf{SD} & 0  &  0   & 0  & \textbf{93}  &  0  &  0  &  6  &  0 \\
			\hline
			\textbf{SU} & 0  &  0 &   0 &   0 & \textbf{106} &   0 &   3  &  0 \\
			\hline
			\textbf{ST} &  1  &  1  &  5  &  0  &  0 & \textbf{604}  & 6  & 3 \\
			\hline
			\textbf{W} & 3  &  2  &  0  &  3   & 2   & 6  &  \textbf{1983 } & 8  \\
			\hline
			\textbf{T} & 7  & 11  & 70   & 2  &  1  & 24  & 48 & \textbf{ 114}  \\
			\hline
			
		\end{tabular}
	\end{table}
	
	\begin{table}[t]
		\centering
		\caption{Confusion Matrix of Baseline Architecture on w-HAR Dataset}
		\label{HAR_Confusion_Matrix_Baseline}
		\begin{tabular}{|c||c|c|c|c|c|c|c|c|}
			\hline
			\multirow{2}{*}{\textbf{True label}} &
			\multicolumn{8}{c|}{\textbf{Second Output Block (Baseline architecture)}} \\
			\cline{2-9}
			& \textbf{J} & \textbf{L} & \textbf{S} & \textbf{SD} & \textbf{SU} & \textbf{ST} & \textbf{W} & \textbf{T}\\
			\hline\hline
			\textbf{J} & \textbf{450}  &  0  &  0   & 1  &  0  &  1   & 2   & 4 \\
			\hline
			\textbf{L} &  0 & \textbf{474}   & 0  &  0   & 0  &  0   & 0  &  0 \\
			\hline
			\textbf{S} & 0  &  0 & \textbf{688}  &  0   & 0  &  8  &  0  &  0 \\
			\hline
			\textbf{SD} &  0  &  0  &  0  & \textbf{94}  &  0  &  0 &   5  &  0 \\
			\hline
			\textbf{SU} & 0  &  0  &  0   & 1  & \textbf{105}  &  0  &  3  & 0 \\
			\hline
			\textbf{ST} &  0  &  1 &   3  &  0  &  0  & \textbf{607}  &  6  &  3 \\
			\hline
			\textbf{W} & 5 & 1  &  0  &  0  &  1   & 6   & \textbf{1986} & 8 \\
			\hline
			\textbf{T} &  4  &  3  & 14  &  0 &   0  & 16 & 18 &  \textbf{222}\\
			\hline
			
		\end{tabular}
	\end{table}

	\begin{table}[t]
	\centering
	\caption{Confusion Matrix of the Output Block Predictor (OBP)}
	\label{OBP_Confusion_Matrix}
	\begin{tabular}{|c||c|c||c|c|}
		\hline
		\multirow{2}{*}{\textbf{True label}} &
		\multicolumn{2}{c||}{\textbf{Opportunity}} &\multicolumn{2}{c|}{\textbf{w-HAR}} \\
		\cline{2-3}\cline{4-5}
		& \textbf{FOB} & \textbf{Baseline} & \textbf{FOB} & \textbf{Baseline} \\
		\hline\hline
		\textbf{FOB} &\textbf{2932 }& 44 & \textbf{4582} & 22 \\
		\hline
		\textbf{Baseline} & 36 & \textbf{119} & 11 & \textbf{125}\\
		\hline
	\end{tabular}
\end{table}

	\begin{table}[t]
		\centering
		\caption{Confusion Matrix of Adaptive Architecture on w-HAR Dataset}
		\label{HAR_Confusion_Matrix_Adaptive}
		\begin{tabular}{|c||c|c|c|c|c|c|c|c|}
			\hline
			\multirow{2}{*}{\textbf{True label}} &
			\multicolumn{8}{c|}{\textbf{Adaptive architecture}} \\
			\cline{2-9}
			& \textbf{J} & \textbf{L} & \textbf{S} & \textbf{SD} & \textbf{SU} & \textbf{ST} & \textbf{W} & \textbf{T}\\
			\hline\hline
			\textbf{J} & \textbf{455}  &  0  &  0  &  0  &  0  &  1  &  1  &  1\\
			\hline
			\textbf{L} & 0 & \textbf{474}  &  0  &  0  &  0  &  0  &  0  &  0\\
			\hline
			\textbf{S} & 0  &  0 & \textbf{688}  &  0  &  0   & 8  &  0  &  0\\
			\hline
			\textbf{SD} & 0 &   0  &  0  & \textbf{97}  &  0  &  0 &  2 & 0\\
			\hline
			\textbf{SU} & 0  &  0  &  0  &  0  & \textbf{106}  &  0  &  3 & 0\\
			\hline
			\textbf{ST} &  1  &  1  &  4  &  0  &  0 &  \textbf{606}  &  6 & 2\\
			\hline
			\textbf{W} &  3  &  2  &  0  &  1  &  2  &  6  &  \textbf{1986} & 7 \\
			\hline
			\textbf{T} & 3 &   3 &  17  &  0  &  0 &  17 & 18 & \textbf{219} \\
			\hline
			
		\end{tabular}
	\end{table}

	\subsection{Performance Evaluation of Multi-output CNN Classifier}
	The performance for each output block of our multi-output CNN classifier is given in Tables \ref{Opportunity_Performance_Table} and \ref{HAR_Performance_Table}. As shown in Table \ref{Opportunity_Performance_Table} for the Opportunity dataset, the FOB has overall accuracy, precision, recall, and weighted F1 score of 87.86\%, 88.54\%, 87.86\% and 87.24\% respectively, whereas; the baseline architecture shows higher overall accuracy, precision, recall, and weighted F1 score of  91.79\%, 91.80\%, 91.79\%, and 91.79\% respectively. The confusion matrices of the output blocks are presented in Table \ref{Opp_Confusion_Matrix}. It shows that the FOB performs poorly in classifying lying activity (63), whereas baseline architecture shows an improved performance (184). Similarly, for the w-HAR dataset, the baseline architecture achieves higher performance than the FOB. As shown in Table  \ref{HAR_Performance_Table}, the FOB achieves an overall accuracy, precision, recall, and weighted F1 score of 95.06\%, 94.87\%, 95.06\% and 94.45\% respectively, whereas; the baseline architecture achieves better accuracy, precision, recall, and weighted F1 score of  97.60\%, 97.57\%, 97.60\%, and 97.55\% respectively. Table \ref{HAR_Confusion_Matrix_FOB} shows that the FOB can classify only 114 transition segments correctly, whereas the baseline architecture shows a better performance while classifying 222 segments correctly as shown in Table \ref{HAR_Confusion_Matrix_Baseline}.
	
		\begin{figure}[t]
		\centering
		\includegraphics[trim={13cm 16.5cm 21cm 5cm},clip, width=\linewidth]{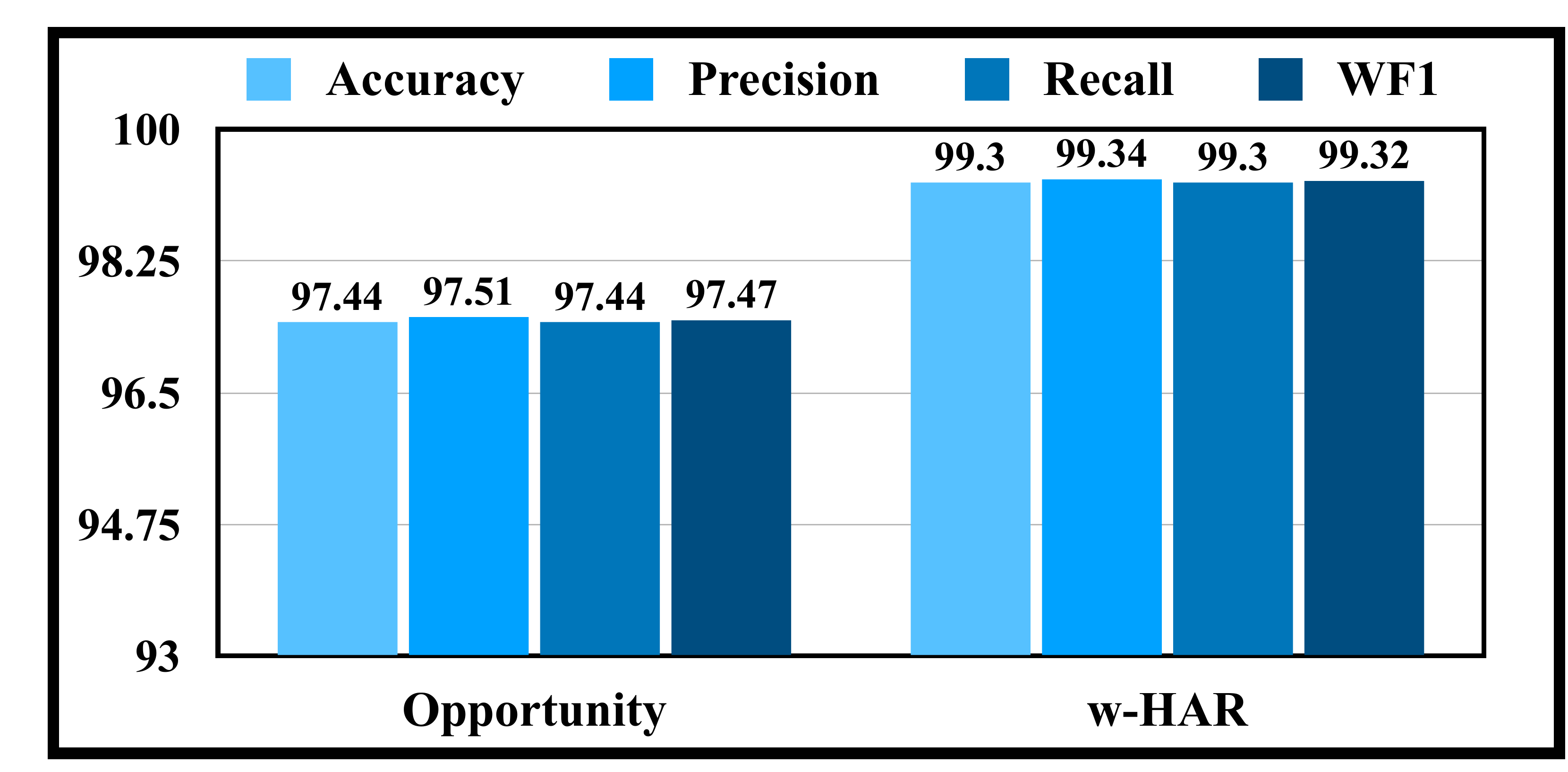}
		\caption{Performance of Output Block Predictor (OBP)}
		\label{OBP_Performance}
	\end{figure}

	\subsection{Performance Evaluation of Output Block Predictor}
	To ensure a better performance of our adaptive architecture, our OBP has to perform better as well.  As shown in Figure \ref{OBP_Performance} for the Opportunity dataset, the OBP has an accuracy, precision, recall, weighted F1 score of 97.44\%, 97.51\%, 97.44\%, 97.47\% respectively. Table \ref{OBP_Confusion_Matrix} shows the corresponding confusion matrix of the OBP in deciding which output block to use for the 3131 test segments of the Opportunity dataset. Similarly, for the w-HAR dataset, the OBP achieves an accuracy, precision, recall, weighted F1 score of 99.30\%, 99.34\%, 99.30\%, and 99.32\% respectively, as shown in Figure \ref{OBP_Performance}. The corresponding confusion matrix for the w-HAR dataset is shown in Table \ref{OBP_Confusion_Matrix}. Both confusion matrix shows that our OBP performs quite well in deciding the required output block to classify the activity segments.
	
	%
	%
	
	\subsection{Performance Evaluation of Adaptive Architecture}
	The performance of the adaptive architecture depends on the decision of OBP. We use the FOB to classify the segments that are predicted as 1 by the OBP. Similarly, the baseline architecture is used to classify the segments that are predicted as 2. The performance of adaptive architecture is determined by the combined performance of the FOB and baseline architecture in classifying the corresponding segments decided by the OBP.
	As shown in Table \ref{Opportunity_Performance_Table} for Opportunity dataset, our adaptive architecture achieves 91.57\% performance for all four metrics- accuracy, precision, recall, and weighted F1 score. It shows that our adaptive architecture achieves very close performance as our baseline architecture while classifying most of the segments (2968) using the FOB. Table \ref{Opp_Confusion_Matrix} shows how the adaptive architecture takes the advantage of both the output blocks.  For example, 63 out of 185 lying activity segment is correctly classified by FOB whereas the baseline architecture can correctly classify 184 of them. And our adaptive architecture can classify 181 of them which is close to the baseline one. Moreover, both our baseline and adaptive architecture outperforms the work \cite{RF-4086Features} and achieves a comparable performance with respect to \cite{DeepConvLSTM, Emilio-like-Survey, I-CNN} as shown in Table \ref{HAR_Performance_Table}. It is to note that the works \cite{RF-4086Features, DeepConvLSTM, Emilio-like-Survey, I-CNN} are designed for fog/cloud platform whereas our solution is designed for wearable platform.
	
	For the w-HAR dataset, our adaptive architecture outperforms our baseline architecture with an accuracy, precision, recall, and weighted F1 score of 97.70\%, 97.69\%, 97.70\%, 97.64\% respectively as shown in Table \ref {HAR_Performance_Table}. Moreover, both our baseline and adaptive architecture outperforms both the baseline and activity-aware classifier used in \cite{Umit_HAR}. As shown in Table \ref{HAR_Confusion_Matrix_Adaptive}, the adaptive architecture can classify 455 out of 458 jump activity whereas the FOB and baseline architecture can classify 445 and 450 of them respectively. This is because there were jump activities that were being classified either by FOB or baseline architecture only. The adaptive architecture uses the best of the two which results in improved performance. Therefore, it proves that the adaptive architecture achieves similar (Opportunity) or better performance (w-HAR) with respect to our baseline architecture while using the FOB to classify most of the segments.

	\begin{table}[t]
		\centering
		\caption{Energy and Memory Consumption Evaluation of the Works on Opportunity Dataset}
		\label{Memory_Power_Consumption_Table_Opportunity}
		\begin{tabular}{|c|c|c|c|c|c|} 
			\hline
			
			\multirow{2}{*}{\textbf{Works}} & \textbf{Classifier} &  \textbf{RAM} &\textbf{Exe.} &   \textbf{Avg. pwr.} & \textbf{Energy} \\
			& \textbf{level} & \textbf{ (Bytes)} &\textbf{time (ms)}& \textbf{(mW)}& \textbf{($\mu$J)} \\
			\hline\hline
			\textbf{\cite{RF-4086Features}} & -  & 60932 & 11722.14 & 16.59 &  194470.31 \\
			\hline
			\textbf{\cite{DeepConvLSTM}} & - & \multicolumn{4}{c|} {Not compatible: RAM overflowed}\\ 
			\hline
			\textbf{\cite{Emilio-like-Survey}} & - & \multicolumn{4}{c|} {Not compatible: RAM overflowed} \\ 
			\hline
			\textbf{\cite{I-CNN}} & - & \multicolumn{4}{c|} {Not compatible: RAM overflowed} \\ 
			\hline\hline
			\multirow{4}{*}{\textbf{[Ours]}}& OBP  & 1120 & 1.61  & 15.26 & 24.56 \\ 
			\cline{2-6} 
			& FOB  & 2688 & 24.57 & 15.25 & 374.69 \\ 
			\cline{2-6}
			& Baseline  & 4264 &  30.25 & 15.22 & 460.41 \\ 
			\cline{2-6}
			& Adaptive & 4264& 26.48 & 15.25 & 403.71\\ 
			\hline		
		\end{tabular}
	\end{table}

	\subsection{Energy and Memory Evaluation on Real Hardware}
	\label{Energy_Memory_Evaluation}
	We evaluate the energy and memory consumption of our proposed architecture including the related works using the EFM32 Giant Gecko microcontroller as mentioned in Section \ref{Wearable_Platform}. For the works \cite{RF-4086Features, Umit_HAR} that uses machine learning approaches, the reported execution time, power, energy, and RAM are for the feature extraction and classification together. For the works using CNN, we evaluate the classification as they automatically extract features during classification. The execution time, power, and energy values presented in the Table \ref{Memory_Power_Consumption_Table_Opportunity} and \ref{Memory_Power_Consumption_Table_HAR} are for one activity segment of data using the 14 MHz clock speed of the microcontroller.
	
	As shown in Table \ref{Memory_Power_Consumption_Table_Opportunity}, for Opportunity dataset the works \cite{DeepConvLSTM, I-CNN, Emilio-like-Survey} using deep CNN encountered RAM overflow and could not be executed on the target hardware. It shows that this kind of solution is only suitable for fog/cloud platforms with higher computational resources. Although the work \cite{RF-4086Features} is designed for fog/cloud platform, it executes on the target hardware with around 60KB of RAM. It takes around 11.72 seconds to extract features and classify an activity segment with 194.47 $\mu$J of energy consumption. On the other hand, our FOB executes with only 2.62 KB of RAM. It takes only 24.57 ms with an energy consumption of 374.69 $\mu$J to classify an activity segment. Our baseline architecture requires higher resources than the FOB as expected. 
	As shown in Table \ref{Memory_Power_Consumption_Table_Opportunity}, the OBP takes only 1.09 KB of RAM to execute. It takes only 1.61 ms to extract 4 statistical features from each segment and classify it with an energy consumption of 24.56 $\mu$J. It shows that the OBP is very lightweight and does not add much overhead to implement our adaptive architecture. To evaluate the execution time and energy of our adaptive architecture, we calculate the average time and energy to classify 3131 test segments either by FOB or baseline architecture based on the decision of our OBP as presented in the confusion matrix of Table \ref{OBP_Confusion_Matrix}. 
	Table  \ref{Memory_Power_Consumption_Table_Opportunity} shows that the adaptive architecture takes only 26.48 ms with an energy consumption of 403.71 $\mu$J which is less than our baseline architecture while providing similar performance.
	
	\begin{table}[t]
		\centering
		\caption{Energy and Memory Consumption Evaluation  of the Works on w-HAR Dataset}
		\label{Memory_Power_Consumption_Table_HAR}
		\begin{tabular}{|c|c|c|c|c|c|} 
			\hline
			
			\multirow{2}{*}{\textbf{Works}} & \textbf{Classifier}  & \textbf{RAM} &\textbf{Exe.} &   \textbf{Avg. pwr.} & \textbf{Energy} \\
			& \textbf{level} &\textbf{(Bytes)} &\textbf{time (ms)}& \textbf{(mW)}& \textbf{($\mu$J)} \\
			\hline\hline
			\multirow{4}{*}{\textbf{\cite{Umit_HAR}}}& Baseline & 9988 & 63.85  & 15.30 & 976.91 \\ 
			\cline{2-6} 
			& Static &  2164 & 31.93 & 15.31 & 488.84 \\ 
			\cline{2-6}
			& Dynamic &  9988 & 85.55 & 15.30 & 1308.92 \\ 
			\cline{2-6}
			& A. aware&  9988 & 65.07 & 15.3 & 995.77 \\ 
			\hline\hline
			\multirow{4}{*}{\textbf{[Ours]}}& OBP & 1128 & 1.96  & 15.23 & 29.86 \\ 
			\cline{2-6} 
			& FOB & 3216 & 26.36 & 15.24 & 401.73 \\ 
			\cline{2-6}
			& Baseline &  4568 & 32.07 & 15.24 & 488.74 \\ 
			\cline{2-6}
			& Adaptive &  4568 & 28.50 & 15.24 & 434.29 \\ 
			\hline
		\end{tabular}
	\end{table}
	
	For the w-HAR dataset, first, we evaluate the baseline classifier of the work in \cite{Umit_HAR}. As shown in Table \ref{Memory_Power_Consumption_Table_HAR}, the baseline classifier takes 63.85 ms to classify a segment with 976.91 $\mu$J of energy. It takes 9.75 KB of RAM to execute. The baseline classifier in \cite{Umit_HAR} involves extracting 120 statistical features from the activity segment and then classify it with a neural network. The activity-aware classifier uses different classifier for static - \textit{Sit (S), Lie (L), Stand (ST)} and dynamic - \textit{Stairs up (SU), Stairs down (SD), Jump (J), Walk (W), Transition (T)} activities.  
	For classifying a static activity, it takes 31.93 ms with 488.84 $\mu$J energy. For the dynamic activities, it consumes higher energy of 1308.92  $\mu$J with a longer execution time of 85.55 ms. The execution time and energy for the activity-aware classifier reported in Table \ref{Memory_Power_Consumption_Table_HAR} is the average time and energy to classify 4740 segments either by the static or dynamic classifier. Out of 4740 segments, the SVM classifier classifies 1810 segments as static and 2930 segments as dynamic as mentioned in \cite{Umit_HAR}. Therefore, the total time and energy for classifying 1810 segments by the static classifier and 2930 segments by dynamic classifier is calculated and summed up. Next, the summation is averaged by 4740 which gives us the average time of 65.07 ms and energy of 995.77 $\mu$J required by the activity-aware classifier. The activity-aware classifier also requires the 9.75 KB of RAM same as the baseline. This RAM is required for calculating the 120 features which is done in both baseline and activity-aware classifier. 
	
	On the other hand, our FOB executes with only 3.14 KB of RAM and takes only 26.36 ms with an energy consumption of 401.73 $\mu$J to classify an activity segment of the w-HAR dataset. As expected, our baseline architecture requires higher resources - 4.46 KB of RAM and 32.07 ms to classify a segment with 488.74 $\mu$J of energy. The OBP takes only 1.10 KB of RAM to execute which takes only 1.96 ms to extract 6 statistical features from each segment and classify it with an energy consumption of 29.86 $\mu$J. Therefore, the OBP takes very minimum resources which ensures minimal overhead to implement our adaptive architecture. To evaluate the execution time and energy of our adaptive architecture, we follow the same procedure as the Opportunity dataset and do it for 4740 segments of the w-HAR dataset. As shown in Table \ref{OBP_Confusion_Matrix}, the OBP decides 4593 and 147 segments to be classified by the FOB and the baseline architecture respectively. Therefore, the summation of total time and energy taken by the OBP, FOB, and baseline architecture is averaged by 4740 which gives us the average time and energy to classify a particular segment by our adaptive architecture. Table  \ref{Memory_Power_Consumption_Table_HAR} shows that the adaptive architecture takes only 28.50 ms with an energy consumption of 434.29 $\mu$J which is less than our baseline architecture while providing better performance.

	\begin{figure}[t]
		\centering
		\includegraphics[trim={13cm 17.5cm 16cm 6.5cm},clip, width=\linewidth]{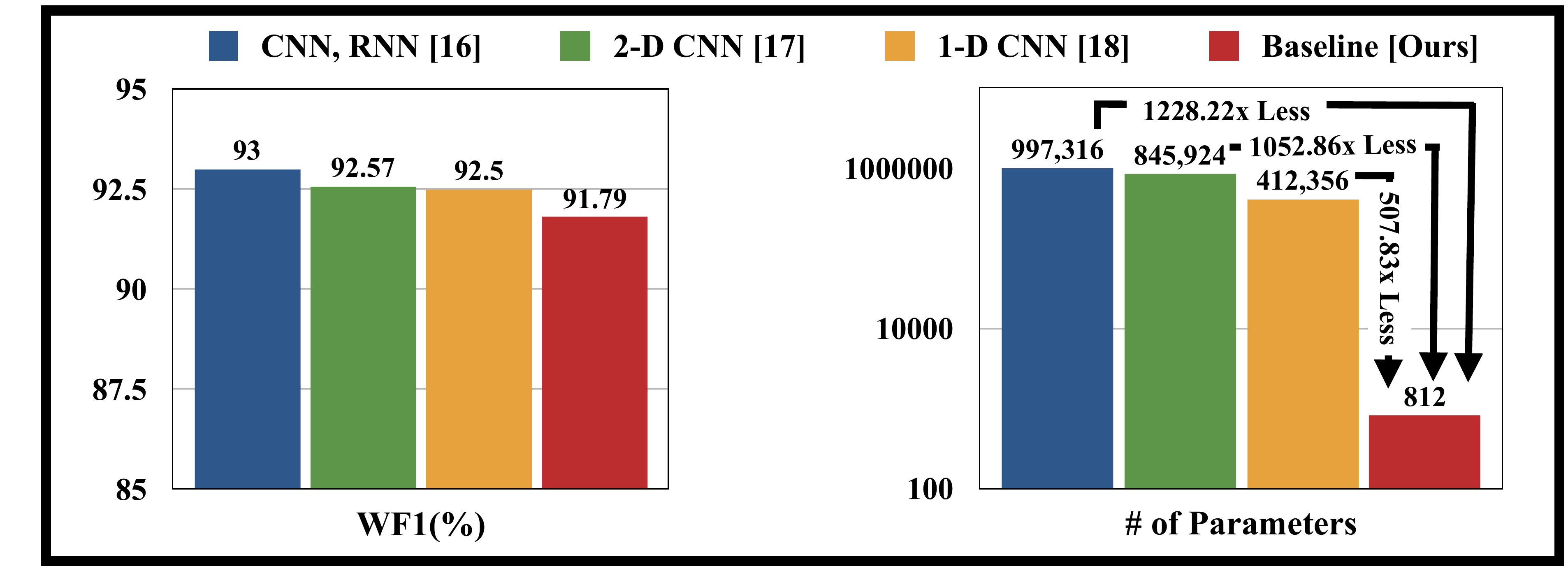}
		\caption{Benchmarking of the deep CNN works on Opportunity dataset}
		\label{Opportunity_F1_Parameters}
	\end{figure}
	
	\begin{figure}[t]
		\centering
		\includegraphics[trim={9cm 15.7cm 12cm 5.5cm},clip, width=\linewidth]{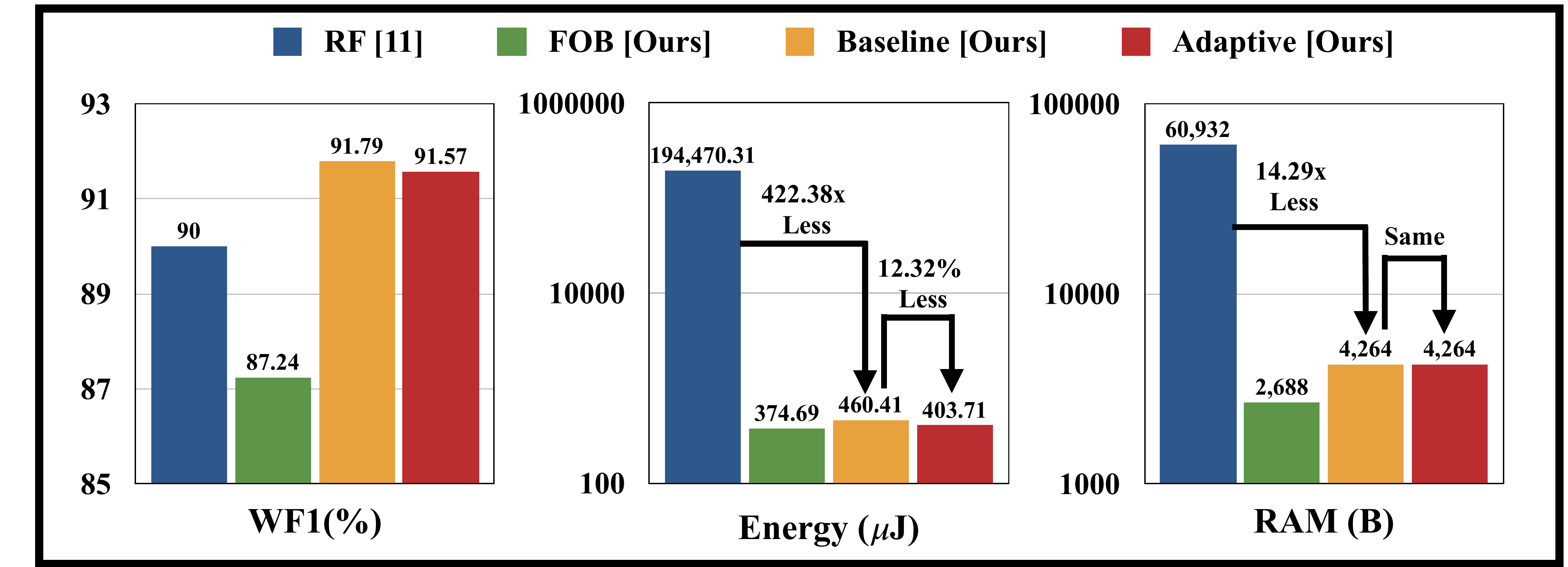}
		\caption{Benchmarking on Opportunity dataset}
		\label{Opportunity_F1_Energy_Memory}
	\end{figure}
	
	\begin{figure}[t]
		\centering
		\includegraphics[trim={9cm 15.4cm 11.5cm 5.5cm},clip, width=\linewidth]{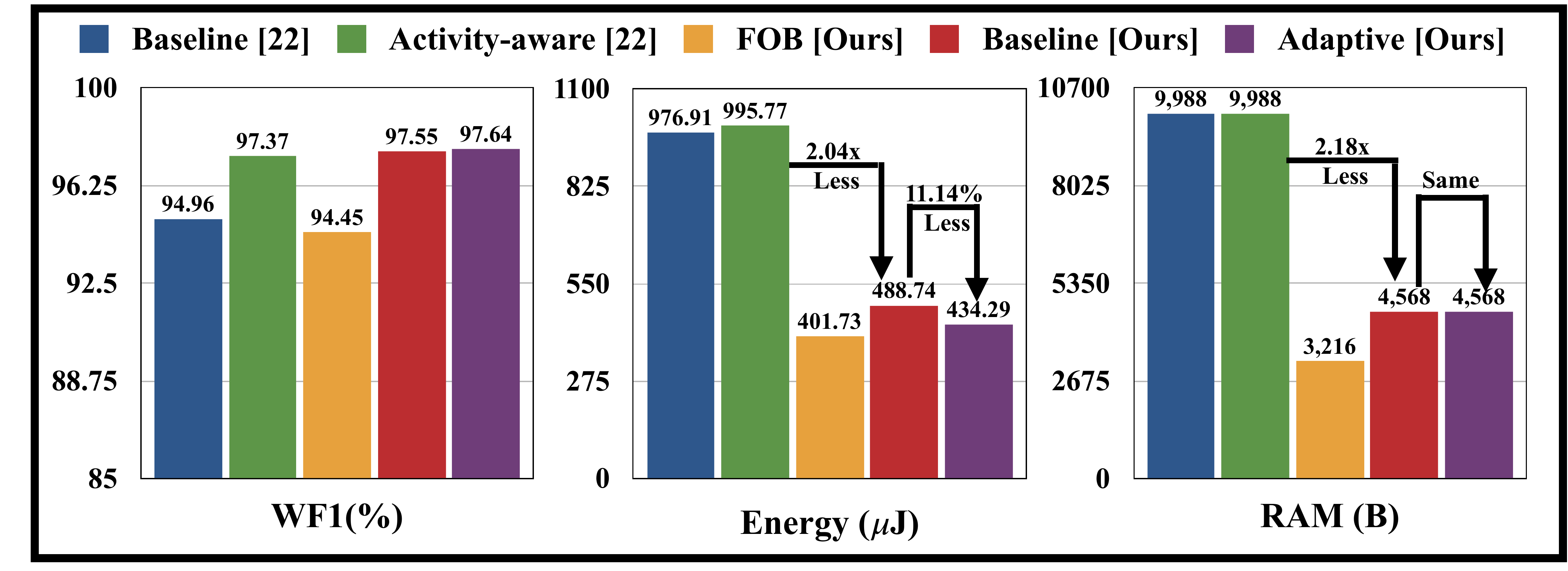}
		\caption{Benchmarking on w-HAR dataset}
		\label{HAR_F1_Energy_Memory}
	\end{figure}

	\subsection{Final Benchmarking}
	
	Finally, we make an overall comparison among the performance of different works along with the computational resources they require. As the deep CNN works \cite{DeepConvLSTM, Emilio-like-Survey, I-CNN} on Opportunity dataset are designed for the fog/cloud platform and do not fit into our target wearable platform, we compare their performance with the network parameter size to give a perspective. As shown in Figure \ref{Opportunity_F1_Parameters}, they achieve higher weighted F1 score of 93\%, 92.57\%, and 92.50\%, whereas; our baseline architecture achieves a comparable weighted F1 score of 91.79\% with 1228.22x, 1052.86x, and 507.83x less parameter size compared to \cite{DeepConvLSTM, Emilio-like-Survey, I-CNN} respectively. Besides, our baseline architecture outperforms the work \cite{RF-4086Features} while consuming 422.38x less energy and 14.29x less RAM as shown in Figure \ref{Opportunity_F1_Energy_Memory}. Moreover, our adaptive architecture achieves similar performance as the baseline while being  12.32\% energy-efficient.
	
	As shown in Figure \ref{HAR_F1_Energy_Memory} for the w-HAR dataset, our baseline architecture outperforms both the baseline (94.96\%) and activity-aware (97.37\%) classifier in \cite{Umit_HAR} with a weighted F1 score of 97.55\% while being 2.04x and 2.18x energy and memory-efficient compared to the activity-aware classifier. Moreover, our adaptive architecture outperforms our baseline while being 11.14\% energy-efficient.
	
	It is to note that, the 12.32\% or 11.14\% energy efficiency achieved by our adaptive architecture over the baseline one is only for 2 layers of convolution. The energy efficiency would be even more if we had deeper CNN architecture with multiple convolution layers. Therefore, our future plan is to investigate the potential of our adaptive CNN architecture for other applications that require multiple convolution layers.

	\section{Conclusion}
	This paper proposes an Adaptive CNN for HAR (AHAR) to develop an energy-efficient solution for low-power edge devices. AHAR uses a novel adaptive architecture that decides which portion of the baseline architecture to be used during the inference phase based on the simple statistical features of the activity segments. Our proposed methodology is validated for classifying locomotion activities from Opportunity and w-HAR datasets. Compared to the fog/cloud computing approaches that use the Opportunity dataset, both our baseline and adaptive architecture shows a comparable weighted F1 score of 91.79\%, 91.57\% respectively. For the w-HAR dataset, both our baseline and adaptive architecture outperforms the state-of-art-work with a weighted F1 score of 97.55\% and 97.64\% respectively. Evaluation on real hardware shows that our baseline architecture is significantly energy-efficient (422.38x less) and memory-efficient (14.29x less) compared to the works on the Opportunity dataset. For the w-HAR dataset, our baseline architecture requires 2.04x less energy and 2.18x less memory compared to the state-of-the-art work. Moreover, experimental results show that our adaptive architecture is 12.32\% (Opportunity) and 11.14\% (w-HAR) energy-efficient than our baseline while providing similar (Opportunity) or better (w-HAR) performance with no significant memory overhead. To the best of our knowledge, we are the first to propose such adaptive CNN architecture for HAR in wearable devices that provides energy efficiency while maintaining performance.
	
	\section{Acknowledgement}
	This work is partially supported by the Graduate Assistance in Areas of National Need (GAANN) award from the United States Department of Education. This paper reflects the views of the authors, not the funding agency.
	
	\bibliographystyle{IEEEtran}
	\bibliography{sample-base2}

\begin{thebibliography}{10}
\providecommand{\url}[1]{#1}
\csname url@samestyle\endcsname
\providecommand{\newblock}{\relax}
\providecommand{\bibinfo}[2]{#2}
\providecommand{\BIBentrySTDinterwordspacing}{\spaceskip=0pt\relax}
\providecommand{\BIBentryALTinterwordstretchfactor}{4}
\providecommand{\BIBentryALTinterwordspacing}{\spaceskip=\fontdimen2\font plus
\BIBentryALTinterwordstretchfactor\fontdimen3\font minus
  \fontdimen4\font\relax}
\providecommand{\BIBforeignlanguage}[2]{{%
\expandafter\ifx\csname l@#1\endcsname\relax
\typeout{** WARNING: IEEEtran.bst: No hyphenation pattern has been}%
\typeout{** loaded for the language `#1'. Using the pattern for}%
\typeout{** the default language instead.}%
\else
\language=\csname l@#1\endcsname
\fi
#2}}
\providecommand{\BIBdecl}{\relax}
\BIBdecl

\bibitem{bort2014measuring}
J.~Bort-Roig, N.~D. Gilson, A.~Puig-Ribera, R.~S. Contreras, and S.~G. Trost,
  ``Measuring and influencing physical activity with smartphone technology: a
  systematic review,'' \emph{Sports medicine}, vol.~44, no.~5, pp. 671--686,
  2014.

\bibitem{bourke2008threshold}
A.~K. Bourke and G.~M. Lyons, ``A threshold-based fall-detection algorithm
  using a bi-axial gyroscope sensor,'' \emph{Medical engineering \& physics},
  vol.~30, no.~1, pp. 84--90, 2008.

\bibitem{parkinson_monitoring}
W.~Maetzler, J.~Klucken, and M.~Horne, ``A clinical view on the development of
  technology-based tools in managing parkinson's disease,'' \emph{Movement
  Disorders}, vol.~31, no.~9, pp. 1263--1271, 2016.

\bibitem{wireless_manik}
M.~{Dautta}, A.~{Jimenez}, K.~K.~H. {Dia}, N.~{Rashid}, M.~A.~A. {Faruque}, and
  P.~{Tseng}, ``Wireless qi-powered, multinodal and multisensory body area
  network for mobile health,'' \emph{IEEE Internet of Things Journal}, pp.
  1--1, 2020.

\bibitem{HHAR}
M.~{Abdel-Basset}, H.~{Hawash}, V.~{Chang}, R.~K. {Chakrabortty}, and
  M.~{Ryan}, ``Deep learning for heterogeneous human activity recognition in
  complex iot applications,'' \emph{IEEE Internet of Things Journal}, pp. 1--1,
  2020.

\bibitem{smartphone_survey}
M.~Shoaib, S.~Bosch, O.~D. Incel, H.~Scholten, and P.~J. Havinga, ``A survey of
  online activity recognition using mobile phones,'' \emph{Sensors}, vol.~15,
  no.~1, pp. 2059--2085, 2015.

\bibitem{lara2012mobile}
O.~D. Lara and M.~A. Labrador, ``A mobile platform for real-time human activity
  recognition,'' in \emph{2012 IEEE consumer communications and networking
  conference (CCNC)}.\hskip 1em plus 0.5em minus 0.4em\relax IEEE, 2012, pp.
  667--671.

\bibitem{smartphone_energy}
L.~M.~S. Morillo, L.~Gonzalez-Abril, J.~A.~O. Ramirez, D.~la~Concepcion, and
  M.~A. Alvarez, ``Low energy physical activity recognition system on
  smartphones,'' \emph{Sensors}, vol.~15, no.~3, pp. 5163--5196, 2015.

\bibitem{wearable_survey}
O.~D. Lara and M.~A. Labrador, ``A survey on human activity recognition using
  wearable sensors,'' \emph{IEEE communications surveys \& tutorials}, vol.~15,
  no.~3, pp. 1192--1209, 2012.

\bibitem{khan2010triaxial}
A.~M. Khan, Y.-K. Lee, S.~Y. Lee, and T.-S. Kim, ``A triaxial
  accelerometer-based physical-activity recognition via augmented-signal
  features and a hierarchical recognizer,'' \emph{IEEE transactions on
  information technology in biomedicine}, vol.~14, no.~5, pp. 1166--1172, 2010.

\bibitem{RF-4086Features}
J.~Zhu, R.~San-Segundo, and J.~M. Pardo, ``Feature extraction for robust
  physical activity recognition,'' \emph{Human-centric Computing and
  Information Sciences}, vol.~7, no.~1, p.~16, 2017.

\bibitem{RF-SVM-KNN}
E.~Fullerton, B.~Heller, and M.~Munoz-Organero, ``Recognizing human activity in
  free-living using multiple body-worn accelerometers,'' \emph{IEEE Sensors
  Journal}, vol.~17, no.~16, pp. 5290--5297, 2017.

\bibitem{Deep-CNN}
W.~Qi, H.~Su, C.~Yang, G.~Ferrigno, E.~De~Momi, and A.~Aliverti, ``A fast and
  robust deep convolutional neural networks for complex human activity
  recognition using smartphone,'' \emph{Sensors}, vol.~19, no.~17, p. 3731,
  2019.

\bibitem{Deep-CNN2}
W.~Jiang and Z.~Yin, ``Human activity recognition using wearable sensors by
  deep convolutional neural networks,'' in \emph{Proceedings of the 23rd ACM
  international conference on Multimedia}, 2015, pp. 1307--1310.

\bibitem{Deep-CNN3}
C.~A. Ronao and S.-B. Cho, ``Human activity recognition with smartphone sensors
  using deep learning neural networks,'' \emph{Expert systems with
  applications}, vol.~59, pp. 235--244, 2016.

\bibitem{DeepConvLSTM}
F.~J. Ord{\'o}{\~n}ez and D.~Roggen, ``Deep convolutional and lstm recurrent
  neural networks for multimodal wearable activity recognition,''
  \emph{Sensors}, vol.~16, no.~1, p. 115, 2016.

\bibitem{Emilio-like-Survey}
E.~Sansano, R.~Montoliu, and {\'O}.~Belmonte~Fern{\'a}ndez, ``A study of deep
  neural networks for human activity recognition,'' \emph{Computational
  Intelligence}, 2020.

\bibitem{I-CNN}
E.~Kim, ``Interpretable and accurate convolutional neural networks for human
  activity recognition,'' \emph{IEEE Transactions on Industrial Informatics},
  2020.

\bibitem{HEAR}
N.~{Rashid}, M.~{Dautta}, P.~{Tseng}, and M.~A. {Al Faruque}, ``Hear:
  Fog-enabled energy-aware online human eating activity recognition,''
  \emph{IEEE Internet of Things Journal}, vol.~8, no.~2, pp. 860--868, 2021.

\bibitem{Hierarchy_Henkel}
F.~{Samie}, L.~{Bauer}, and J.~{Henkel}, ``Hierarchical classification for
  constrained iot devices: A case study on human activity recognition,''
  \emph{IEEE Internet of Things Journal}, vol.~7, no.~9, pp. 8287--8295, 2020.

\bibitem{EDGE}
H.~Li, K.~Ota, and M.~Dong, ``Learning iot in edge: Deep learning for the
  internet of things with edge computing,'' \emph{IEEE network}, vol.~32,
  no.~1, pp. 96--101, 2018.

\bibitem{Umit_HAR}
G.~Bhat, Y.~Tuncel, S.~An, H.~G. Lee, and U.~Y. Ogras, ``An ultra-low energy
  human activity recognition accelerator for wearable health applications,''
  \emph{ACM Transactions on Embedded Computing Systems (TECS)}, vol.~18,
  no.~5s, pp. 1--22, 2019.

\bibitem{EMBC_2020}
N.~{Rashid} and M.~A. {Al Faruque}, ``Energy-efficient real-time myocardial
  infarction detection on wearable devices,'' in \emph{2020 42nd Annual
  International Conference of the IEEE Engineering in Medicine Biology Society
  (EMBC)}, 2020, pp. 4648--4651.

\bibitem{CNN}
Y.~LeCun, Y.~Bengio, and G.~Hinton, ``Deep learning,'' \emph{nature}, vol. 521,
  no. 7553, pp. 436--444, 2015.

\bibitem{rashid2021feature}
N.~Rashid, L.~Chen, M.~Dautta, A.~Jimenez, P.~Tseng, and M.~A. Al~Faruque,
  ``Feature augmented hybrid cnn for stress recognition using wrist-based
  photoplethysmography sensor,'' in \emph{2021 43rd Annual International
  Conference of the IEEE Engineering in Medicine \& Biology Society
  (EMBC)}.\hskip 1em plus 0.5em minus 0.4em\relax IEEE, 2021, pp. 2374--2377.

\bibitem{panda2016conditional}
P.~Panda, A.~Sengupta, and K.~Roy, ``Conditional deep learning for
  energy-efficient and enhanced pattern recognition,'' in \emph{2016 Design,
  Automation \& Test in Europe Conference \& Exhibition (DATE)}.\hskip 1em plus
  0.5em minus 0.4em\relax IEEE, 2016, pp. 475--480.

\bibitem{scardapane2020should}
S.~Scardapane, M.~Scarpiniti, E.~Baccarelli, and A.~Uncini, ``Why should we add
  early exits to neural networks?'' \emph{arXiv preprint arXiv:2004.12814},
  2020.

\bibitem{w_HAR}
G.~Bhat, N.~Tran, H.~Shill, and U.~Y. Ogras, ``w-har: An activity recognition
  dataset and framework using low-power wearable devices,'' \emph{Sensors},
  vol.~20, no.~18, p. 5356, 2020.

\bibitem{OPPORTUNITY}
D.~Roggen, A.~Calatroni, M.~Rossi, T.~Holleczek, K.~F{\"o}rster,
  G.~Tr{\"o}ster, P.~Lukowicz, D.~Bannach, G.~Pirkl, A.~Ferscha \emph{et~al.},
  ``Collecting complex activity datasets in highly rich networked sensor
  environments,'' in \emph{2010 Seventh international conference on networked
  sensing systems (INSS)}.\hskip 1em plus 0.5em minus 0.4em\relax IEEE, 2010,
  pp. 233--240.

\bibitem{lu2019neural}
Q.~Lu, W.~Jiang, X.~Xu, Y.~Shi, and J.~Hu, ``On neural architecture search for
  resource-constrained hardware platforms,'' \emph{arXiv preprint
  arXiv:1911.00105}, 2019.

\bibitem{liu2018darts}
H.~Liu, K.~Simonyan, and Y.~Yang, ``Darts: Differentiable architecture
  search,'' \emph{arXiv preprint arXiv:1806.09055}, 2018.

\bibitem{odema2021energy}
M.~Odema, N.~Rashid, and M.~A. Al~Faruque, ``Energy-aware design methodology
  for myocardial infarction detection on low-power wearable devices,'' in
  \emph{2021 26th Asia and South Pacific Design Automation Conference
  (ASP-DAC)}.\hskip 1em plus 0.5em minus 0.4em\relax IEEE, 2021, pp. 621--626.

\bibitem{EExNAS}
M.~Odema, N.~Rashid, and M.~A.~A. Faruque, ``Eexnas: Early-exit neural
  architecture search solutions for low-power wearable devices,'' in \emph{2021
  IEEE/ACM International Symposium on Low Power Electronics and Design
  (ISLPED)}, 2021, pp. 1--6.

\bibitem{odema2021lens}
M.~Odema, N.~Rashid, B.~U. Demirel, and M.~A.~A. Faruque, ``Lens: Layer
  distribution enabled neural architecture search in edge-cloud hierarchies,''
  in \emph{2021 58th ACM/IEEE Design Automation Conference (DAC)}, 2021.

\bibitem{cai2020once}
\BIBentryALTinterwordspacing
H.~Cai, C.~Gan, T.~Wang, Z.~Zhang, and S.~Han, ``Once for all: Train one
  network and specialize it for efficient deployment,'' in \emph{International
  Conference on Learning Representations}, 2020. [Online]. Available:
  \url{https://arxiv.org/pdf/1908.09791.pdf}
\BIBentrySTDinterwordspacing

\bibitem{courbariaux2016binarized}
M.~Courbariaux, I.~Hubara, D.~Soudry, R.~El-Yaniv, and Y.~Bengio, ``Binarized
  neural networks: Training deep neural networks with weights and activations
  constrained to+ 1 or-1,'' \emph{arXiv preprint arXiv:1602.02830}, 2016.

\bibitem{wang2019haq}
K.~Wang, Z.~Liu, Y.~Lin, J.~Lin, and S.~Han, ``Haq: Hardware-aware automated
  quantization with mixed precision,'' in \emph{Proceedings of the IEEE
  conference on computer vision and pattern recognition}, 2019, pp. 8612--8620.

\bibitem{hinton2015distilling}
G.~Hinton, O.~Vinyals, and J.~Dean, ``Distilling the knowledge in a neural
  network,'' \emph{arXiv preprint arXiv:1503.02531}, 2015.

\bibitem{chen2016eyeriss}
Y.-H. Chen, T.~Krishna, J.~S. Emer, and V.~Sze, ``Eyeriss: An energy-efficient
  reconfigurable accelerator for deep convolutional neural networks,''
  \emph{IEEE journal of solid-state circuits}, vol.~52, no.~1, pp. 127--138,
  2016.

\bibitem{chen2019eyeriss}
Y.-H. Chen, T.-J. Yang, J.~Emer, and V.~Sze, ``Eyeriss v2: A flexible
  accelerator for emerging deep neural networks on mobile devices,'' \emph{IEEE
  Journal on Emerging and Selected Topics in Circuits and Systems}, vol.~9,
  no.~2, pp. 292--308, 2019.

\bibitem{EFMGG}
\BIBentryALTinterwordspacing
S.~Labs. (2021) Efm32™ giant gecko 32-bit microcontroller. [Online].
  Available: \url{https://www.silabs.com/mcu/32-bit/efm32-giant-gecko}
\BIBentrySTDinterwordspacing

\end{thebibliography}
	
\end{document}